\pdfoutput=1
\documentclass[10pt,twocolumn,letterpaper]{article}

\usepackage[pagenumbers]{cvpr}      

\usepackage{graphicx}
\usepackage{amsmath}
\usepackage{amssymb}
\usepackage{booktabs}
\usepackage{tabularx}
\usepackage{subcaption}

\usepackage[pagebackref,breaklinks,colorlinks]{hyperref}

\usepackage{float}
\usepackage[capitalize]{cleveref}
\crefname{section}{Sec.}{Secs.}
\Crefname{section}{Section}{Sections}
\Crefname{table}{Table}{Tables}
\crefname{table}{Tab.}{Tabs.}

\begin{document}

\date{January 2023}
\title{Behavioural Cloning in VizDoom}
\author{
  Spick, Ryan$^*$\\
  \texttt{ryan.spick@sony.com}
  \and
  Bradley, Tim$^*$\\
  \texttt{timothy.bradley@sony.com}
  \and
  Raina, Ayush\\
  \texttt{ayush.raina@sony.com}
  \and
  Amadori, Pierluigi Vito\\
  \texttt{pierluigi.vito.amadori@sony.com}
  \and
  Moss, Guy\\
  \texttt{guy.moss@sony.com}
}
\maketitle
\def\thefootnote{*}\footnotetext{Equal Contribution}

\begin{abstract}
This paper describes methods for training autonomous agents to play the game ``Doom 2'' through Imitation Learning (IL) using only pixel data as input. We also explore how Reinforcement Learning (RL) compares to IL for humanness by comparing camera movement and trajectory data. Through behavioural cloning, we examine the ability of individual models to learn varying behavioural traits. We attempt to mimic the behaviour of real players with different play styles, and find we can train agents that behave aggressively, passively, or simply more human-like than traditional AIs. We propose these methods of introducing more depth and human-like behaviour to agents in video games. The trained IL agents perform on par with the average players in our dataset, whilst outperforming the worst players. While performance was not as strong as common RL approaches, it provides much stronger human-like behavioural traits to the agent.

\end{abstract}

\section{Introduction}
\label{sec:intro}

Deep learning has become popular in recent years because of its ability to learn and make predictions from complex data with minimal human intervention, the combinations of increased computing resources and better designed algorithms and architectures have led to state-of-the-art deep learning applications in many fields \cite{openai_chatgpt, chen2021decision,tan2019efficientnet}. One of the major advantages of deep learning is its ability to learn from raw data, without the need for human involvement. This makes it a powerful tool for handling complex data types such as images \cite{tan2019efficientnet}, audio \cite{purwins2019deep}, and natural language \cite{openai_chatgpt}. Deep Neural Networks (DNN) are composed of multiple layers of interconnected neurons that can learn complex representations of the data through a process of iterative training on large datasets \cite{goodfellow2016deep}.

In recent years, DNNs have shown promising results in the field of behavioural cloning (BC) \cite{pearce2022counter, ho2016generative}. BC is a form of \textit{Imitation Learning} (IL), where we train an artificial ``\textit{agent}'' to mimic actions from an observable state of expert data \cite{torabi2018behavioral}. Agents are trained using a number of historical states, be they image frames or other data, and their corresponding actions. The learning is performed by using the final frame's associated action as the "target", this target being passed to some loss function. The loss function will reinforce the observed frame's predicted action, doing this over an extremely large dataset will achieve an agent that can predict the best action to take at any one given set of input image frames \cite{osa2018algorithmic}. 

Reinforcement learning (RL) is a type of machine learning that involves an agent learning from experience by directly interacting with the environments and receiving feedback on the quality of its actions in the form of reward signals \cite{Sutton1998}. The goal of the agent is to learn a policy that maximises its cumulative reward over time.

The main difference in these approaches resides in how agents learn their policies. While IL agents only rely on previous expert demonstrations, RL agents also need to have access to the reward signals they receive when interacting with the environment itself \cite{Sutton1998}. RL has shown that interaction with the environment and access to rewards signals are very beneficial to autonomous agents training, as its policies often lead in state-of-the-art performance. However, extracting reward signals is a rather demanding process, as it requires the development of special APIs \cite{Kempka2016ViZDoom} to access engine-specific data.  

Although the use of APIs to access additional engine data is extremely beneficial when leveraged correctly for training high performing policies \cite{remonda2021formula}, accessing it requires either previously built support for gathering this data, access to the source code, memory profiling, or estimation from the image sequence, which in itself is just a reconfiguration of an end to end task of learning a policy directly from image data. For the vast majority of games, at the time of writing, access to engine data is not directly available. By focusing in on image streams, and data that can be reliably inferred from natural images, such as segmentation maps \cite{zhao2023fast}, and depth buffers \cite{piccinelli2023idisc}, which have become more viable in computer games as the graphical fidelity of modern games gain a greater level of realism \cite{shafaei2016play}

In this paper we compare approaches from VizDoom \cite{Kempka2016ViZDoom, Wydmuch2019ViZdoom} agents trained using solely IL and RL approaches. We discuss the advantages and disadvantages of both methods and outline their effective results in the VizDoom training environment. In line with the above, the proposed approaches make use of RGB data in combination with additional data such as depth buffers and segmentation. We access this data directly from the engine owing to the low graphical fidelity of the environment.  The main goal is to provide an IL-based approach that can learn a robust policy from an RGB stream alongside the collected key inputs, all of which could be obtained by recording someone playing any type of game.

The rest of the paper is structured as follows: 
\textit{Section \ref{sec:related}} reviews the prior literature that relates to this work;
\textit{Section \ref{sec:background}} outlines the theory and tools used in this work;
\textit{Section \ref{sec:method}} describes how the data was collected, experiments were configured, and ablation studies were done;
\textit{Section \ref{sec:Evaluation}} presents the results of the experiments from the previous section;
\textit{Section \ref{sec:RL}} compares the behaviour of IL and RL agents, and presents a more in-depth analysis of movement and camera control patterns;
Lastly, results are discussed in \textit{Section \ref{sec:Discussion}} where conclusions are drawn.
\section{Related Work}
\label{sec:related}
\subsection{Learning to Play Games}

Using Neural Network (NN) based approaches to control games is a relatively new idea. Early works in RL used dynamic programming, genetic algorithms, or decision tree methods \cite{kaelbling1996reinforcement}, and often focused on simple games of choice \cite{littman2001friend}. 

From 2009 to 2012, Togelius \textit{et al.} \cite{togelius2013mario} ran the \textit{``Mario AI Championship''}. For the 2009 competition \cite{togelius2010}, participants aimed at designing bots who could complete as many levels as possible of an open source version of \textit{Super Mario Brothers}. The majority of entries used hand-coded search functions and only two entries proposed NN-based approaches (one RL and one IL). The NN-based approaches succeeded in completing only three out of the forty generated levels, and placed near the bottom of the results board. The last year the competition, the organisers included a ``Turing Test'' track for the competition \cite{shaker2013turing}, in which humans would vote on the believability of the various competitors. The only NN-based approach out of the three non-human competitors was soundly defeated.

Shortly thereafter Mnih \textit{et al.} \cite{mnih2013playing} developed \textit{Deep Q-Networks} to play a number of Atari games, triggering a resurgence in popularity of DNNs in RL.

Recently, focus in the field has shifted towards transformers, with the decision transformer created by Chen \textit{et al.} \cite{chen2021decision} being of particular note, and Large Language Models (LLMs) \cite{openai_chatgpt}, owing in part to their ability to better handle temporal data. Kwon \textit{et al.} \cite{kwon2023reward} use an LLM to design reward functions for a number of simple games, and show promising results compared to the more traditional baseline. Lee \textit{et al.} \cite{lee2022multi} use a generalised decision transformer to learn to play multiple games simultaneously.

Besides RL, IL has also gained increasing attention as a promising approach for enabling agents to learn complex behaviours from expert demonstrations, particularly in the fields of robotics \cite{haldar2023teach} and video games \cite{Wydmuch2019ViZdoom}. Recently, DeepMind used IL to create agents that are capable of interacting with humans in a game environment \cite{team2021creating} and then fine-tune those agents using RL in a human-in-the-loop paradigm \cite{abramson2022improving} with excellent results.

Closest to our work is that of Pearce \& Zhu \cite{pearce2022counter}. The authors used BC to train their agents to play \textit{Counter Strike Global Offensive} (CSGO) with a combination of EfficientNet \cite{tan2019efficientnet}, a convolutional LSTM \cite{shi2015convolutional}, and a series of dense layers, achieving very good results. To do so, they exploited a critic loss based on agents performance to train the agents and used a one-hot encoding to model mouse movements. On the other hand, in our approach we only use visual feed to train our agents, as it eases implementation and improves generalisability, and formulated mouse movements as a regression problem, as one-hot encoding proved to be not effective in VizDoom.

\section{Background}
\label{sec:background}

This section offers a brief overview of autonomous agents, and describes the scenario our agents attempt to learn.

\subsection{Autonomous Agents}
Here we define an autonomous agent as a system that can control a game or device without the need for human interference, or having being specifically programmed to control that game or device. The aims of creating intelligent agents that play games in an intuitive manner are fourfold: 

Firstly, increased realism and challenge through the development of more realistic and competent opponents or allies for players. 

Secondly, enabling agents to react and learn from new obstacles and strategies without intervention from a game developer.

Thirdly, increasing the diversity of agent behaviour through the incorporation of learnt strategies, which we can achieve by preconditioning our agents on particular players' or player groups' data.

Lastly, aiding developers by removing the need to code complex logic, and providing agents who can be used to automate quality control and testing steps.

Broadly speaking approaches to creating autonomous agents fall into one or more of the following categories: \textit{Imitation Learning}, in which an agent is trained in a supervised manner to imitate the actions of an expert performing a task; \textit{Reinforcement Learning}, wherein an agent is trained on a provided signal from its environment, which represents its performance at its given task; finally \textit{Genetic Programming} and \textit{Evolutionary Programming}, which are similar but independent approaches to having an agent randomly mutate its behaviour or structure until it meets some performance benchmark in a given task. 
Genetic and Evolutionary programming fall outside the scope this paper, but we refer the interested reader to a review by Hua \textit{et al.} \cite{hua2021survey}, which outlines theory and application of evolutionary algorithms.

\subsubsection{Imitation Learning}

Imitation learning algorithms learn by example, whether that is a robotic arm controller learning to copy the actions of a human \cite{johns2021coarse} to complete a given task, an autopilot learning to fly a plane based on the actions of a pilot with feedback from the planes instruments \cite{sammut1992learning}, or an agent in a game learning to copy the trajectories and actions of human players \cite{pearce2022counter}.

Formally, Imitation learning aims to identify a policy $\hat{\pi_{\theta}}$, parameterised by 
$\theta$, that approximates the policy of an expert $\pi$. Here, the policy is a function that outputs an action, $\mathbf{a}$, given a current state $\mathbf{s}$ \cite{pmlr-v9-ross10a}:
\begin{align}
\mathbf{a} = \pi(\mathbf{s}).
\label{eq:policy}
\end{align}

The policy is then learned by minimising some cost function $J$ by minimising a loss, $\mathcal{L}$, providing the cost of following a policy $\hat{\pi}$, in a given state, $s$, sampled from a distribution of expert data $P(a|\pi)$:
\begin{align}
J(\theta) = \mathop{\arg \min}\limits_{\theta} \mathbb{E}_{s\sim P(s|\pi)} \left[ \mathcal{L} \left(s, \hat{\pi}_{\theta}\left(s\right)\right) \right].
\end{align}

Though in inverse reinforcement learning we would aim to maximise the reward, one can simply substitute the loss function in the above equation with the negative of the reward function. In BC the problem is reduced to a supervised learning task \cite{pmlr-v15-ross11a}, which we can state as:
\begin{align}
J(\theta) = \mathop{\arg \min}\limits_{\theta} \sum\limits_{t=1}\limits^{N}  \mathcal{L} \left(\pi\left(s_t\right), \hat{\pi}_{\theta}\left(s_t\right)\right),
\end{align}
where $N$ is the number of expert state observations in our dataset. Substituting in from Equation \ref{eq:policy} we arrive at:
\begin{align}
J(\theta) = \mathop{\arg \min}\limits_{\theta} \sum\limits_{t=1}\limits^{N}  \mathcal{L} \left(\mathbf{a}_t, \hat{\mathbf{a}}_t\right),
\end{align}
where $\mathbf{a}$ is the action from the expert data and $\hat{\mathbf{a}_t}$ is the example produced by the trained policy.

\subsection{Environment}
VizDoom \cite{Kempka2016ViZDoom} is a python-based platform for interacting with the 1994 First Person Shooter (FPS) game, Doom2. Doom2 offers both single player modes, where the player fights against computer controlled bots and monsters, and multiplayer modes where players can compete with each other over a network. In recent years, VizDoom has grown popular in the field of computer vision and machine visual learning, also supported by its use for a competition on autonomous agents \cite{Wydmuch2019ViZdoom}.

VizDoom grants direct access to a wide variety of data such as: RGB rendered scene, depth information, segmentation map, top-down map, telemetry, kill counts, and other game statistics. Thanks to its versatility, it has been incorporated into popular IL and RL tools such as SampleFactory \cite{petrenko2020sf} and EnvPool \cite{envpool}.

In this paper, we focus our studies on the multiplayer deathmatch mode. The environment is a custom-designed map, see Fig. \ref{fig:trajectory_all}, where a central square region with four pillars is surrounded by several corridors and chambers leading to it. The centre of the map has a lava pool that leads to a loss of health if entered. Health packs and ammunition spawn at random locations on the map, while few high value armour items spawn in the centre of the map. The agents are spawned (start from) at the exterior regions of the maps.

\section{Methodology} 
\label{sec:method}

\subsection{Data Collection and Individual Data}

To collect data we built two applications. A server to host games, collect statistics, and save replay files, and a client application that allowed players to create and join lobbies, set some in-game options (such as resolution, character colour, etc), and launch the game locally. This was necessary as ViZDoom doesn't currently support recording gameplay against inbuilt bots, so we needed to gather human vs human data.

Our setup allows us to collect data for multiple maps and rule sets quickly, as we have a viewpoint stored for each player in a given match. The replays are stored as binary LMP files, which are very lightweight and editable to an extent. 

For each match, we also store information about player performance and match rules in an SQL table. The \textit{Player} table includes fields such as number of deaths, number of ``frags'', whether they won the match, the ID number of the match they were in, and a back reference to the \textit{User} table, which is simply a list of all unique players, which can be pivoted on. The \textit{Match} table contains information about individual matches, such as which ViZDoom configuration file was used, which map was played on, the time the match was played, the number of players in the match, the duration of the match, and the replay filename in which the match was saved.

Data were collected across 10 different players over 41 matches, with 6 players playing the majority of the games in the dataset. The variables remained the same during every deathmatch that was played, recording the entire match. Each individual's data was extracted from the replay, saving their actions and game-frame output at a 640x480 resolution. The playback recording speed was set to capture 35 frames a second, therefore a $\sim$3-minute gameplay session produced $\sim$5700 frames per player, after removing $\sim$600 frames to account for match start-up, and the time players spent waiting to respawn. The complete dataset comprised of 95 individual player trajectories, equating to $\sim$541,000 frame-action pairs. 

The actions players could take, and that we subsequently stored were: \texttt{TURN\_LEFT\_RIGHT\_DELTA} and \texttt{LOOK\_UP\_DOWN\_DELTA} which corresponded to mouse movement input in the x and y axes, respectively, and were used to orient the view of the player-character and are stored as floating point values; \texttt{ATTACK} which is a binary variable corresponding to left-click input from the mouse and causes the player-character to fire their weapon; and \texttt{MOVE\_FORWARD}, \texttt{MOVE\_BACKWARD}, \texttt{MOVE\_LEFT}, and \texttt{MOVE\_RIGHT}, which are binary variables mapped to the "W", "A", "S", and "D" keys respectively and are used to move the player-character in the direction indicated by the variable name without reorienting them.

We collected data from over 41 multiplayer deathmatches of Doom2, with varying amounts of players in each game. There were different combinations of players during the various sessions that we ran. Although the dataset contains 10 players, we only explore six of the most played user's data in this paper, owing to insufficient data from the other four.

\begin{figure}[ht] 
\begin{subfigure}{0.25\textwidth}
\includegraphics[width=\linewidth]{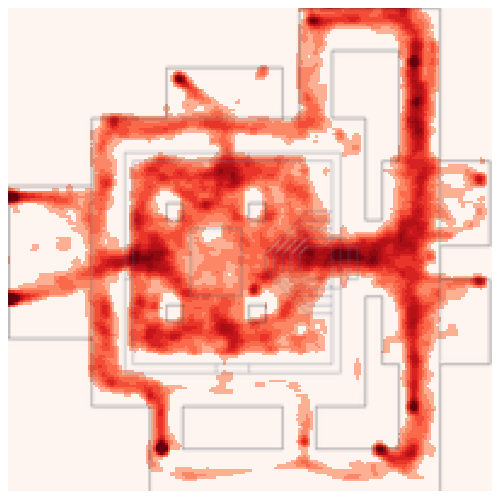}
\caption{Player FadedHeater} \label{fig:a}
\end{subfigure}\hspace*{\fill}
\begin{subfigure}{0.25\textwidth}
\includegraphics[width=\linewidth]{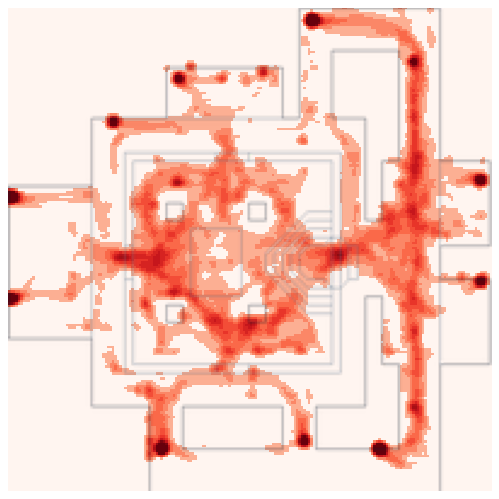}
\caption{Player HospitableKiller} \label{fig:b}
\end{subfigure}

\medskip
\begin{subfigure}{0.25\textwidth}
\includegraphics[width=\linewidth]{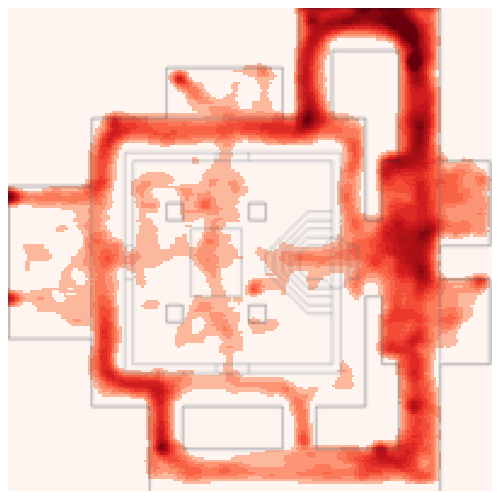}
\caption{Player LeanCeiling} \label{fig:c}
\end{subfigure}\hspace*{\fill}
\begin{subfigure}{0.25\textwidth}
\includegraphics[width=\linewidth]{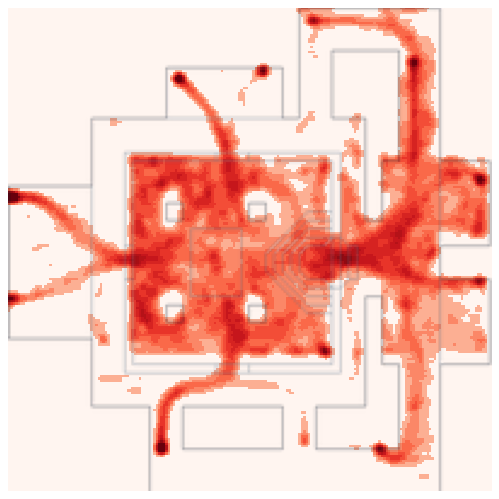}
\caption{Player PointlessSolitaire} \label{fig:d}
\end{subfigure}
\begin{subfigure}{0.25\textwidth}

\medskip
\includegraphics[width=\linewidth]{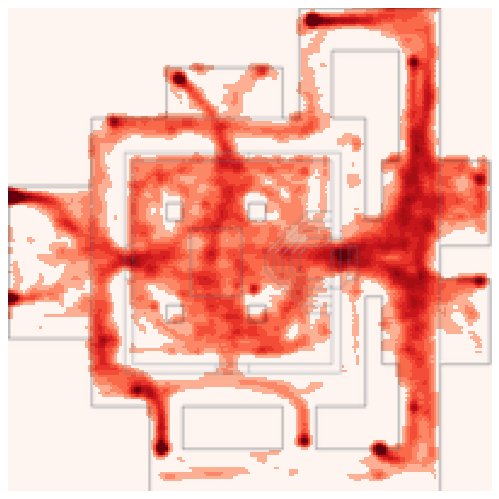}
\caption{Player NebulousFellow} \label{fig:e}
\end{subfigure}\hspace*{\fill}
\begin{subfigure}{0.25\textwidth}
\includegraphics[width=\linewidth]{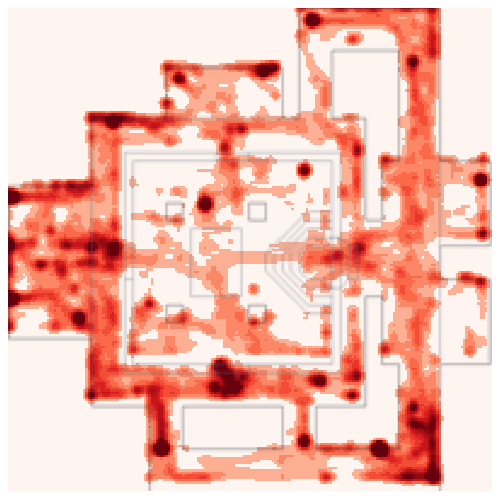}
\caption{Player WastefulTandem} \label{fig:f}
\end{subfigure}

\caption{Figure of multiple trajectories showing each individual player's movement in the deathmatch environment. Each player's username has been anonymised into a random name, these will be referenced when discussing individual player's metrics and agent training.} 
\label{fig:individual_trajectories}
\end{figure}

Fig. \ref{fig:individual_trajectories} shows each individual trajectory of the six players for which we had the most data in the deathmatch collection. The arena itself can be seen outlined in black under the red trajectory paths. It is apparent that there are completely different play styles between each player, with some players opting to stay around the outer edge, whereas some players run straight for the centre of the map. Such largely varied behaviour leads to the ability to train models on each collection of user data, with agents exhibiting similar trajectory patterns to the ones of the experts they trained from. We further expand on this in Section \ref{sec:playerTrainedAgents}.

\begin{table}[ht]
\begin{tabular}{lrrrr}
\textbf{Player}                         & \multicolumn{1}{l}{\textbf{Kills}} & \multicolumn{1}{l}{\textbf{Deaths}} & \multicolumn{1}{l}{\textbf{K/D}} & \multicolumn{1}{l}{\textbf{Win Rate}} \\ \hline
\multicolumn{1}{l|}{PointlessSolitaire} & 20.05                              & 11.14                               & 1.80                                  & 61.9\%                                \\
\multicolumn{1}{l|}{LeanCeiling}        & 15.14                              & 7.89                                & 1.92                                  & 29.7\%                                \\
\multicolumn{1}{l|}{NebulousFellow}     & 15.78                              & 12.86                               & 1.23                                  & 27.0\%                                \\
\multicolumn{1}{l|}{FadedHeater}        & 14.84                              & 11.68                               & 1.27                                  & 12.0\%                                \\
\multicolumn{1}{l|}{HospitableKiller}   & 7.66                               & 15.34                               & 0.50                                  & 0.0\%                                 \\
\multicolumn{1}{l|}{WastefulTandem}     & 5.03                               & 10.31                               & 0.52                                  & 0.0\%                                
\end{tabular}
\caption{The table collects the average performance of the six players who played the most. Throughout the paper, we use the win rate as metric to determine a player's skill level.}
\label{tab:player_data}
\end{table}

\subsection{Ablation/Setup}
In this section we discuss the network setup, the loss functions used for training, and introduce a novel method of frame skipping that increases how far back in time the network can ingest data.

The network consisted of three parts; an initial feature extractor, a Convolutional LSTM, and a fully connected output-head for each action. The initial input was passed through multiple layers of 2D convolutions, using stride to reduce the size of the input, whilst we increased the number of filters with depth. The temporal element of the data was unchanged in this layer. These features were then passed through a convolutional LSTM which extracted features on a temporal level. Lastly, there was a separate un-shared output head for each action and mouse axis, these layers utilised dropout at each full connection. The mouse had no activation function to ensure a continuous regressive output. The action keys had a sigmoid activation applied as they were a binary choice.

The input to our models is a sequence of multimodal frames comprising of RGB, depth, and label information. The RGB is the screen pixel data, the depth map is the distance information and the label is a 2D segmentation map with values ranging from 1-255 for any objects (such as ammo, health packs, etc.). All were captured directly from the players perspective with a $256px \times 192px$ resolution. When building the final input frames, we layered the images on top of each into a single 5-channel frame.

An important part of this work was removing all action, or auxiliary data, and just relying on the observed frames. We did experiment using RGB data alone, and while the agent performed well in navigation, it lacked the basic gameplay loop of finding and shooting other bots. The overall performance gain of the trained agent from the added depth and label data was around $100\%$, in terms of its kill-death ratio.

When modelling the actions, we assumed navigation movements to be binary values and mouse movements to be continuous values. We did not normalise or limit the mouse values. We averaged the action label $l$ for each frame sequence of length $N$ over the next $L$ frames, as follows
\begin{align*}
    l_{t} &= \sum\limits_{n=1}\limits^{L} \frac{l_{t+n}}{L},
\end{align*}
where $l$ is the label at the time step $t$.

We found that averaging the targets helped with both regularisation and stability of the network, as the target label for similar ranges of frame sequences would have a smoother transition, compared to using a single non-averaged label from one of the next frames. For the results in this paper, we set $L=2$.

The mouse loss function followed a similar style to MSE, however, a weighting was placed on the sign prediction of the mouse through the loss. We set the sign mask to $0.33$ for an incorrect sign prediction, and to $1.0$  for a correct sign prediction. The signed-MSE was calculated as follows:
\begin{align*}
SignMask &= (prediction) * (label) > 0 \\
SignMask &= (SignMask + 0.5) / 1.5 \\
MouseLoss &= MSE(prediction,label) / SignMask.
\end{align*}

We tested mouse movement performance when using standard MSE and signed-MSE. Signed-MSE lead to a much more human-like movement pattern. Its importance was apparent at movements close to $0$ when the prediction was either side of $0$ by a small amount, i.e., $\sim 0.05$ or $\sim -0.05$. Here, a standard MSE would produce low loss values even if the sign of the prediction was wrong.

The loss for the key presses and navigation movements, e.g., forward, left, right and attack, was calculated using binary cross entropy. Finally, the two losses were combined and propagated as normal.

Our dataset was quite imbalanced, with the positive values of some actions such as move-forward, far outweighing the negative. Penalties were used to inhibit loss values of those actions that occurred more frequently than others in the dataset. A value of between $0.1$ and $2$ was used to multiply against the corresponding loss component of the action or mouse movement. These values are discussed in more detail in the parameter sweep Section \ref{sec:sweep}, with the specific value shown in Table \ref{tab:params}. We also counter the data imbalance by employing action balancing. We iteratively swap between activated action keys and non-activated keys in a batch, and we can ensure every batch has an equal balance of labels, where randomly chosen actions such as Attack, or Move Right, are $1$ or $0$.

Learning rate warm-up was a simple way of stabilising the early stages of training. This helped with IL training, since a "bad'' set of initial frames and actions would leave the network stuck in some local minima. The warm-up was performed over the first $500$ epochs using a linear function, with the learning rate $\eta$ returning to normal fixed value $\eta \textsuperscript{\textit{i}}$ for all epochs following the $500$-th epoch, as follows
\begin{align*}
\eta = \eta \textsuperscript{\textit{i}} * min((epoch/500),1).
\end{align*}

Frame skipping was a highly effective method of increasing the capacity of the network's historical view of frames, without increasing the input length. Given the frame rate of $35$ Hz, an input of $N=15$ without frame skipping would provide less than half a second worth of frames per prediction, causing serious limitations in the model's capabilities.

Standard frame skipping would follow a linear pattern for history skipping such as taking every \textit{n} frames. In this paper, we introduce a novel exponential frame skipping approach, such that important frames close to time \textit{t} would be denser than those at time \textit{t \textsuperscript f}. The sequence is formed using an "exponent frame offset'' $\lambda_{o}$ as a hyperparameter of the network; for each frame of the sequence we calculate its offset $f_o$ from the time-step $t$ in the sequence using the following formula:
\begin{align*}
f_o(i;\lambda_o) = (i**\lambda_o),
\end{align*}
where \textit{i} is the step in the sequence. Here $1 \leq\lambda_{o}\leq 1.5$, where $1$ indicates no frame skipping. Our image sequence, of length $N$, used to predict an action for timestep $t+1$, is thus:
\begin{align*}
    F_{t}, F_{t-f_o(1)}, F_{t-f_o(2)}, ..., F_{t-f_o(N-1)},
\end{align*}
where $F_{t}$ is the 5-channel stacked frame we defined above. 


\section{Evaluation}
\label{sec:Evaluation}
\subsection{Objective Metrics}
\label{sec:Objective}
Evaluation in IL and other classification problems is usually performed via a validation set and a training set, with the validation set used to calculate the network performance on unseen data.

In our model, although we use a validation set to monitor training, the main evaluation of network performance was carried out by taking checkpoints of the network and running these on an actual doom deathmatch against the default bot script in the game. We used an aggregate of 10 games with each game lasting one minute. Our reasoning for not using validation loss was due to how the IL predictions were often ''incorrect" but could still lead to a solution to the current state the agent was in. 

We collected and used the damage, kills and deaths of the agent averaged over these games to determine performance, with damage as the main factor of performance. We had explored a combination of the 3 variables, but it became apparent that damage correlated to kills anyway, and that high deaths didn't necessarily mean the agent was performing worse than a lower death agent, but that it oftentimes was navigating to more populated areas resulting in higher deaths.

Given how long each of these performance evaluations takes, we would only perform this calculation every 250 epochs. The best-performing model would then be based on the results of the evaluation instead of using the validation loss.

\subsection{Parameter Sweeps}
\label{sec:sweep}
Searching for the optimal parameters was key to finding the best-performing agent, with the majority of parameters combinations leading to sub-optimal-performing agents. This section will discuss which parameters we chose to explore. 

Our initial parameter sweeps won't be outlined in detail but were used to determine which parameters should be frozen, and which should be explored in more detail. At the end of the first sweep, we took all of the models and weighted their parameters based on the performance of the model.

At the end of the initial sweep, we calculated the ranking of each model's performance and its network parameters. These were used to reduce and fix some variables for the final exploration.

The performance of the model was calculated using our outlined objective function in section \ref{sec:Objective}, which calculates damage done for our agent over a $10$-game period of $1$ minute per game.

The batch size was inversely adjusted against the frame length, to ensure we would have the maximum amount of data processed on the hardware at any one run.

The second and final sweep took place with fixed parameters for the image size at $256px \times 192px$, we also reduced the sequence length to $N \in [10, 15]$. Lastly, the penalisation value for each of the actions was changed to mirror the exact same value for similar actions, such as left-right movement and mouse movement

While the first sweep showed that it favoured smaller models, we did not end up changing the exploration in this area, as we believed that the reason it was biased towards smaller models was the training duration was not long enough to fully exploit the additional network capacity. We, therefore, decided to increase the duration of training by twice as long for the final exploration.

The target label was calculated using $1$ of $3$ functions, targeting one random frame,  targeting average frames, and targeting one next frame, with the first $2$ using a random range from $1$ - $5$. These functions were all present in the parameter sweep, with the averaging function leading to the best performance. Averaging was done on all actions from the frame after the last frame in the sequence, with a determined range of $2$.

The exact parameters used in all of the models proceeding this section can be seen in the tables below. Table \ref{tab:penalize} shows the discovered values we used to penalise the loss values for specific actions, this helped with actions that were massively over-represented in the data and also assisted those actions which were more sensitive to incorrect prediction such as moving left or right. Table \ref{tab:params} shows the overall optimal parameters of the network, with the size and depth of each neural layer. 

\begin{table}[hb]
\begin{center}
\begin{tabular}{l|l}
\hline
Action & Value \\
\hline
Attack & 1.25 \\
Move Right & 1.73 \\
Move Left & 1.73 \\
Move Forward & 0.54 \\
Move Backward & 1.94 \\
Mouse Left/Right & 0.45 \\
Mouse Up/Down & 0.45 \\
\hline
\end{tabular}
\caption{Penalising weights for action losses. The weights were multiplied against the corresponding action's loss to promote/penalise more correct/incorrect actions, respectively. }
\label{tab:penalize}
\end{center}
\end{table}

\begin{table}[hb]
\begin{center}
\begin{tabular}{l|r}
\hline
Parameter           & \multicolumn{1}{l}{Value}              \\
\hline
CNN Depth           & 5                                      \\
CNN Layer           & \multicolumn{1}{l}{$74 *  (2^ { depth-1 })$} \\
Conv LSTM Depth     & 4                                      \\
Conv LSTM Layer     & \multicolumn{1}{l}{$9 + (depth*2)$}      \\
MLP Depth           & 5                                      \\
MLP Layer           & \multicolumn{1}{l}{$1984 / (2^{ depth-1})$} \\
Learning Rate       & 0.0002                                 \\
Frame Length        & 15                                     \\
Frame Skip Exponent & 1.22                                   \\
Target Method       & \multicolumn{1}{l}{Average Frames}     \\
Target Range        & 2              \\                       
\hline
\end{tabular}
\caption{Final hyperparameters of the network.}
\label{tab:params}
\end{center}
\end{table}

\subsection{Main Results} 
\label{mainresults}
In this section, we discuss the best-performing agent, which was trained on data from the top 3 players. We also explore an agent trained on data from the bottom 3 players. We show that the best-performing players produce a far better agent than the data from the worst-performing players, we then explore their movement trajectories and in-game agent performance statistics.

\begin{figure}[t!]
    \centering
    \begin{subfigure}[b]{0.5\linewidth}
    \centering
    \frame{\includegraphics[width=0.9\linewidth]{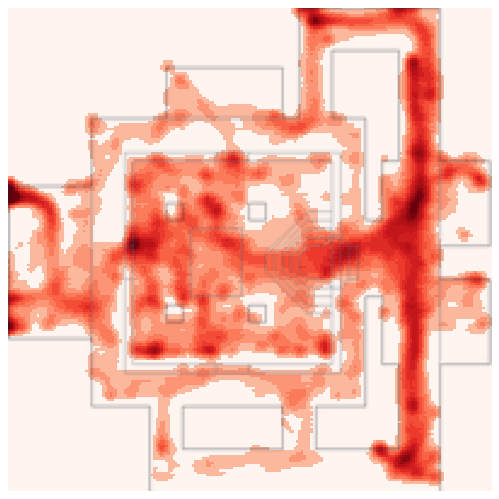}}
    \caption{}
    \label{fig:trajectory_all}
    \end{subfigure}%
    \begin{subfigure}[b]{0.5\linewidth}
    \centering
    \frame{\includegraphics[width=0.9\linewidth]{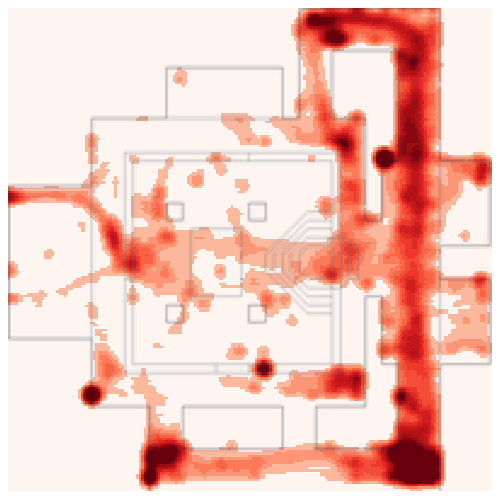}}
    \caption{}
    \label{fig:trajectory_bottom_3}
    \end{subfigure}%
    \caption{a) Agent trajectory heatmap, from a model trained on the top 3 agent's data. b) Agent heatmap from the bottom 3 players combined training} 
\end{figure}

Averaged over eight games the agent trained on the top three players scored the highest damage and kills across all of our runs. Table \ref{tab:agent_perf} describes the agent's performance during these test matches. Whilst there is little difference between individual performance from the top three players, there is a fairly large difference in the performance of the top three players combined together during training, with roughly a $20\%$ improvement in kills and damage over the agent trained on all the players' data. The agent also produced some of the cleanest trajectory patterns, with its trajectory being very balanced across the entire map, with a large portion of the agent's time spent in the map's "hot spots" where most of the action occurred. This can be seen in Fig. \ref{fig:trajectory_all}.

When observing the agent play against other default bots, the navigation was the smoothest of all of the trained agents, with almost no time spent trapped against walls or other collidable objects, which occurred sometimes in the other agent's inference. 

Although we recorded deaths, usually the number of deaths were heavily skewed by the pathing of the agent and the decision to enter the middle of the map over staying around the edges. Overall, the performance when training on the top 3 players surpassed training on individual players, and training on the entire dataset.

When we compare the model trained on the top three players, to the data in Table \ref{tab:agent_perf} of the bottom-3 player's agent we can see a clear divide in performance through the kills and damage statistic. Although the bottom 3 model performs better on the number of deaths, this is systematic of its performance on navigation. The model would often get stuck on objects, and spend much time in the corridors, and when it finally did navigate to areas with other bots, the reduced agent performance led to less damage and kills. 

The trajectory of the bottom 3 player-trained-agent can be seen in Fig. \ref{fig:trajectory_bottom_3}, it is apparent that there are very large differences between the best-trained agent, and the performance of the agent trained on the bottom 3 players. The obvious differences are the exploration levels of the bottom-3 agent. We can also see that the agent's hotspots on the trajectory are far higher, indicating that it would often be in a small area, either stuck on an object or moving back and forth.

Overall the best-performing agent does extremely well in both its in-game performance of targeting and shooting other bots, and also in its ability to navigate around the arena, which when compared to the bottom-3 trained agent, has a much smoother and human-like movement pattern.

\begin{table}[ht]
\begin{tabular}{lrrr}
\hline
                                                & \multicolumn{1}{l}{\textbf{Kills}} & \multicolumn{1}{l}{\textbf{Damage}} & \multicolumn{1}{l}{\textbf{Deaths}} \\ \hline
\multicolumn{1}{l|}{Agent (Top 3)}              & 11                               & 1462                                 & 11.2                                 \\
\multicolumn{1}{l|}{Agent (PointlessSolitaire)} & 9.2                               & 1294                                 & 10.9                                   \\
\multicolumn{1}{l|}{Agent (NebulousFellow)}     & 9.3                              & 1200                                 & 10.2                                 \\
\multicolumn{1}{l|}{Agent (LeanCeiling)}        & 7.8                               & 1134                                 & 7.6                                 \\
\multicolumn{1}{l|}{Agent (Bottom 3)}           & 5.1              & 762             & 8.7     \\       \hline
\end{tabular}
\caption{Agent performance with individual and grouped player training.}
\label{tab:agent_perf}

\end{table}

\subsection{Player Trained Agents}
\label{sec:playerTrainedAgents}
An important part of this study was how agents' behaviour varied when trained only on one type of player data. We, therefore, opted to pick player data based on individual performance and trajectories, see Fig.  \ref{fig:individual_trajectories}. The trajectory maps provide clear insights on the player's game understanding, i.e., higher performing players aim to control specific areas of the map and spend a higher portion of their time in these areas.

\begin{figure}[ht] 
\begin{subfigure}{0.5\textwidth}
\includegraphics[width=\linewidth]{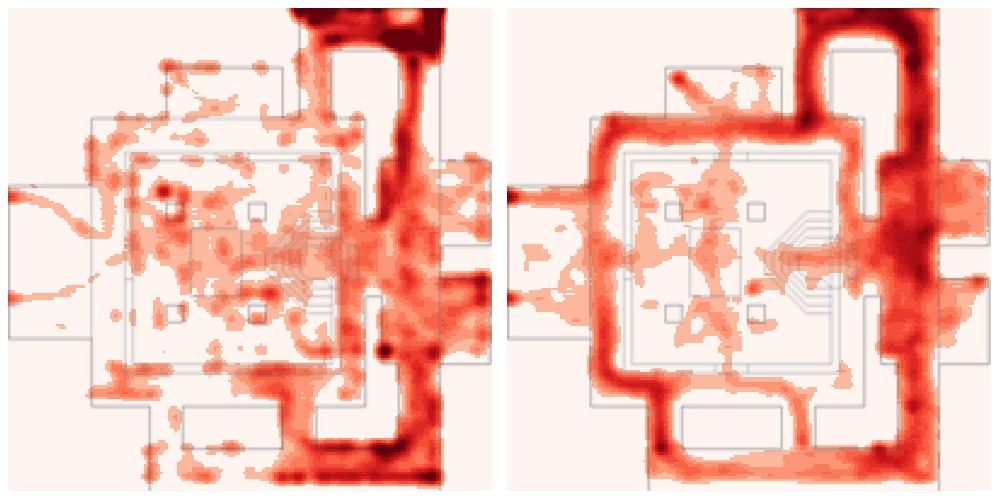}
\caption{Comparison of agent heatmap (left) trained from LeanCeiling (right)}
\end{subfigure}
\begin{subfigure}{0.5\textwidth}
\includegraphics[width=\linewidth]{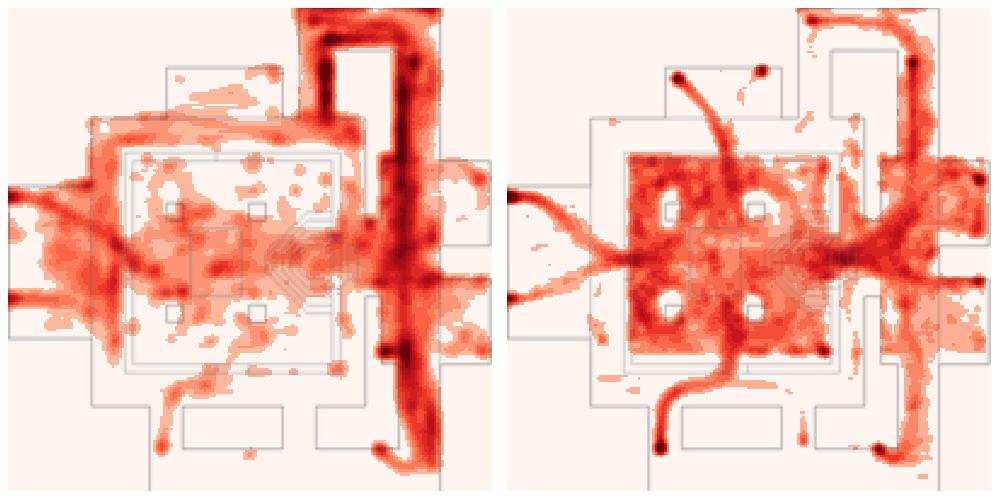}
\caption{Comparison of agent heatmap (left) trained from PointlessSolitaire (right)} 
\end{subfigure}
\begin{subfigure}{0.5\textwidth}
\includegraphics[width=\linewidth]{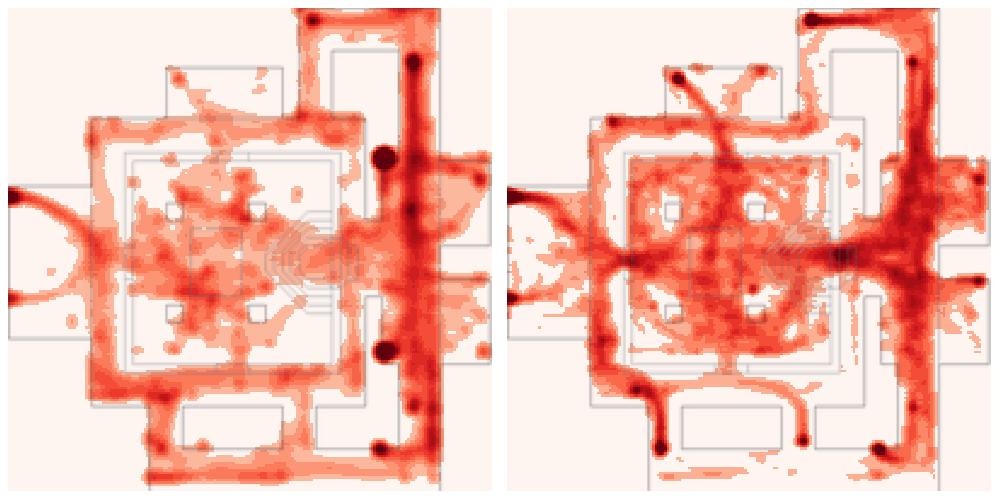}
\caption{Comparison of agent heatmap (left) trained from NebulousFellow (right)} 
\end{subfigure}

\caption{Single-player trained agent's trajectories. Trajectories are from a single game with 6 bot players to most closely match the data captured from the player deathmatches. }\label{fig:bot_vs_player}
\end{figure}

The heatmaps of the three trained agents can be seen in Fig. \ref{fig:bot_vs_player}, these were chosen as the three best-performing players in the dataset by win rate. These trajectories were taken over a 60-minute game against 6 "bots"controlled by the VizDoom's default AI. We used longer game times to more closely resemble the length of game time on a per-player basis used in the player trajectories. Long gameplays were used exclusively to explore the trajectories.

Looking at the three model's resulting exploration data, we can clearly see most of the map is being explored. There are large concentrations of movement around the upper walls, this can be described by the agent getting "stuck" in these areas until it eventually finds a way out. When observing the agent, we found that it would navigate very well, until it found itself in situations that did not exist in the player's datasets, e.g., getting stuck against a wall.

When we compare the agents to their derived training data we start to see patterns that emerge. For LeanCeiling's data, the agent spends a lot less time in the centre of the map, much like the original player's behaviour, it favours the right-hand side and the corners around this region. Interestingly, there appears to be a correlation between an agent that spends more time on the outer map and the lower amount of deaths the agent may have. This could be due to the increased amount of activity in the centre of the map, with the majority of deaths occurring here.

The second model PointlessSolitaire shows a slightly different movement pattern, with more observed time spent in the centre of the map. It does however share similarities to the first model, with the corridors on the right side of the map being traversed more heavily. The original player's data shows a pattern of "rushing" to the centre, which we can observe similar patterns in the model's behaviour. 

Out of all of the agents, NebulousFellow's agent explores the map the most, this can be directly related to their training data, which has a lot more activity in the outer edges of the map, but also spends a lot of time in the centre. 

We categorised these agents into 3 behaviours or personalities; agent LeanCeiling which prefers far safer gameplay, spending less time in the chaotic centre leading to far fewer deaths overall, but also less damage leading to lower perceived performance.
Agent Pointlesssolitaire, which plays an aggressive game with the highest damage done, but also a much higher average death rate. Lastly, NebulousFellow, who can be described as a balanced mix of the other two agents.

Overall the idea of training on individual players' data clearly leads to unique performing agents, with varying implicit ``behaviours''. There are slight variations in the performances of each model, with a much wider variety of movement patterns. 

\section{Comparison to RL Bots}
\label{sec:RL}
This section introduces the agents trained from scratch
using RL and compares their performance with both human data and imitation agents.

\subsection{RL Setup}
For the RL process, every VizDoom game episode is considered as a Markov Decision
Process where a tuple $\langle S, A, P, R \rangle$ is defined for State, Action, Probability, and Reward is defined. We maintain the same definitions from Sample Factory \cite{petrenko2020sf}. The State definition combines visual and game information where the visual information is a stack of 4 RGB image frames of an agent’s view point and the game information is the numerical values visible during the game (e.g., armour, kills, etc). We modify the Action definition by adding 21 discrete new actions for ``Vertical aim'' and remove the ``Weapon selection'' head as the agents only have access to a shotgun, this was to match the IL action space as closely as possible.

The deathmatch scenario in VizDoom is modelled as a multi-entity setup where a self-play population-based training methodology is employed \cite{pbt2019}. The training environment is composed of several instances of randomly initialised policy networks competing with one another. Asynchronous Proximal Policy Optimisation (APPO) \cite{petrenko2020sf} is used to train each of the policy networks. The policies that perform the worst have a fraction of their weight overwritten by the best policies. One episode can last several minutes hence multiple processes execute experience collection episodes in parallel. For population based training we instantiate 4 policies and trained them for 2.5 billion frames each as recommended in previous work \cite{petrenko2020sf}. The policies achieve high performing behaviour and we further visually observe intelligent behaviour for map navigation and attacking mechanics.

\subsection{Behavioural Analysis}
This section contains data analysis and visualisation of the behavioural data in VizDoom. Particularly this work focuses on agents' spatial and camera movements. A spatial heatmap is used to evaluate the spatial movements of the agent in subsection \ref{sm}. The heatmap shows the locations on a map visited by the agent and hence illustrate their preferences and temporal aspects of their movement. The camera movements are compared analytically in subsection \ref{aim}. For both analyses the trained RL agent is used to generate 2 million frames of inference data and action and movement data are recorded for analysis. 

\subsubsection{Spatial movement} 
\label{sm}
\begin{figure}[ht]
    \centering
    \includegraphics[width=0.9\linewidth]{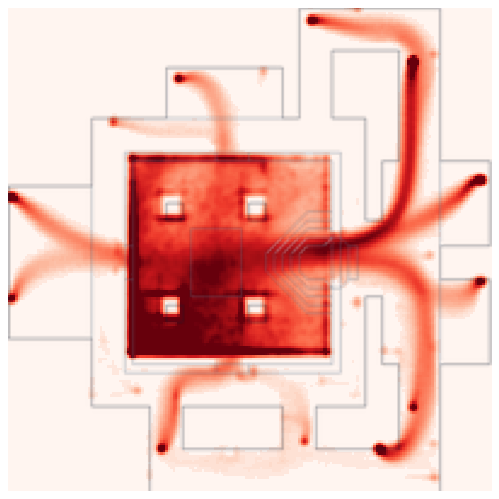}
    \caption{Spatial heatmap of the RL agent's behaviour} \label{fig:rl_trajectory_all}
\end{figure}

Movement data from the RL agent is visualised in the heatmap in Figure \ref{fig:rl_trajectory_all}. The solid lines represent the walls in the environment. The pixel colour in the heatmap represent the frequency of occurrence of the agent in that spatial x, y location of the map and bright colors in adjacent regions capture the movement preference of the agents. The formulation is identical to heatmaps from section \ref{mainresults}. Locations with an occurrence value below a threshold have been masked out and appear transparent to improve visibility and comprehension of the heatmap.

It can be observed from Figure \ref{fig:rl_trajectory_all} that the agents have learned to navigate through the corridors as they turn through corners and avoid walls. Further two main qualitative observations can be made from the heatmap. First, the agents distinctively head towards the centre of the map from their spawn points at the edges of the map. The agents also run over the lava pool in the centre to pick up armor items making them harder to kill before heading towards the edges of the central region. Second, there are bright regions around the pillars meaning that the agents go around these pillars. On visual analysis it can be confirmed that agents use the pillars as cover from enemy fire as a means of making themselves harder to shoot at.

Further these heatmaps can be compared with human and imitation agent behaviours from Figure \ref{fig:bot_vs_player}. The RL agents share similar behaviour characteristics to the high performing humans as in Figure \ref{fig:bot_vs_player} part (b) and (c). Both RL agents and top humans move towards the centre from their spawn point and spend time around the pillars, which is distinctly different from the other humans in the dataset. This illustrates that independently the RL agents have identified high performing spatial behaviour. This analysis confirms that the spatial movements of the RL agents are similar to high performing human behaviour.

\subsubsection{Camera movement} 
\label{aim}
\begin{figure}[ht]
    \centering
    \includegraphics[width=0.9\linewidth]{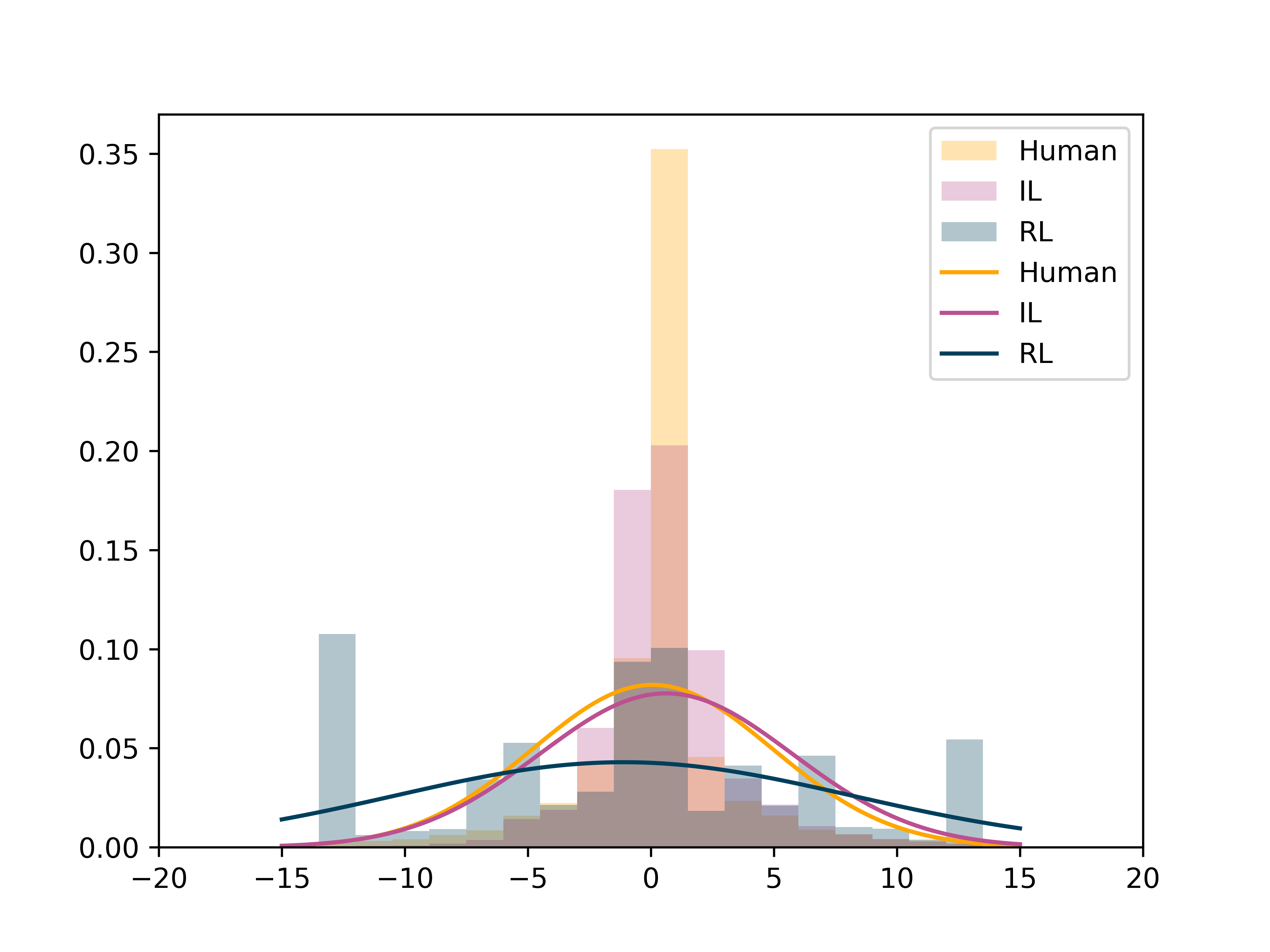}
    \includegraphics[width=0.9\linewidth]{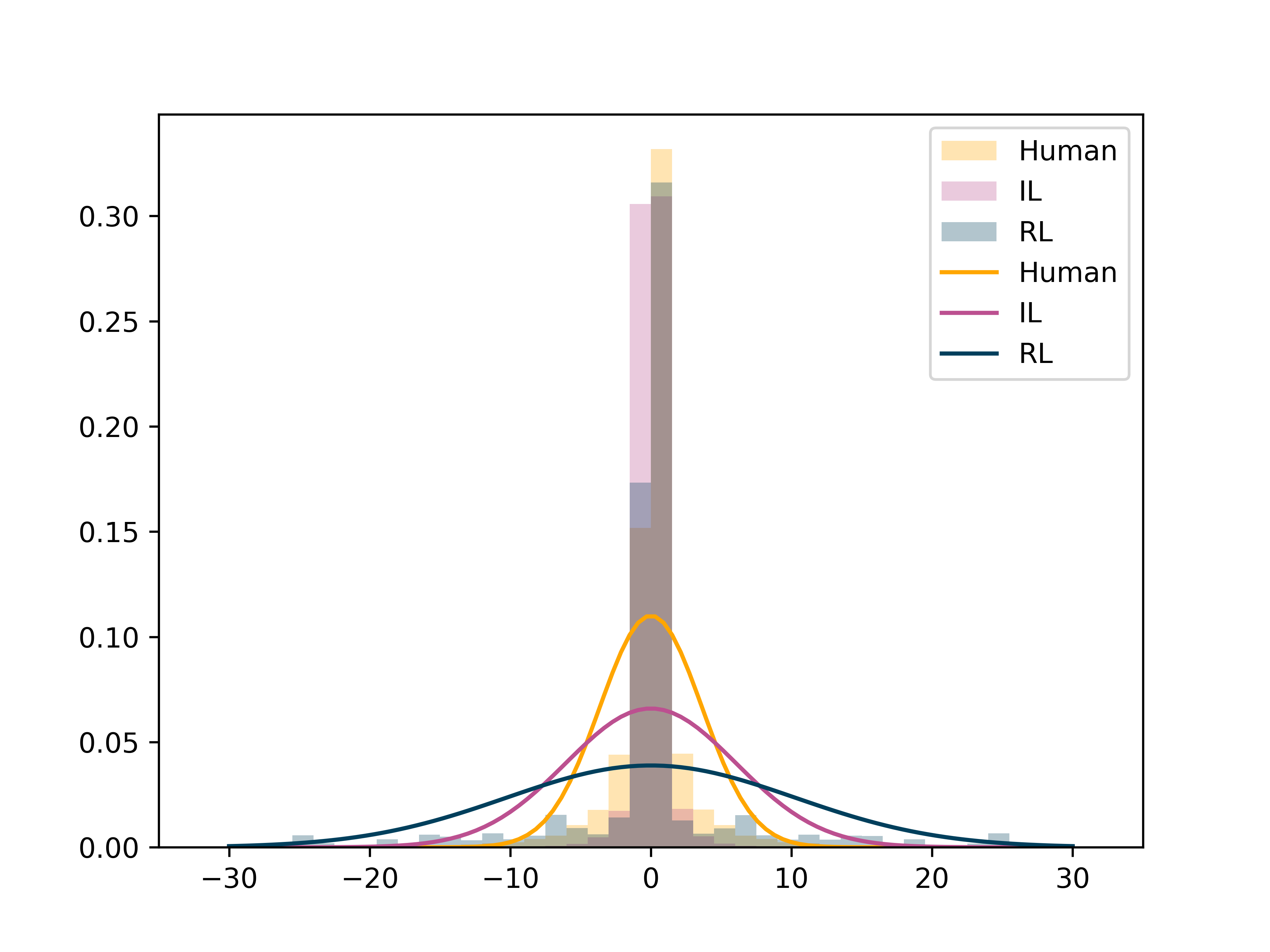}
    \includegraphics[width=0.9\linewidth]{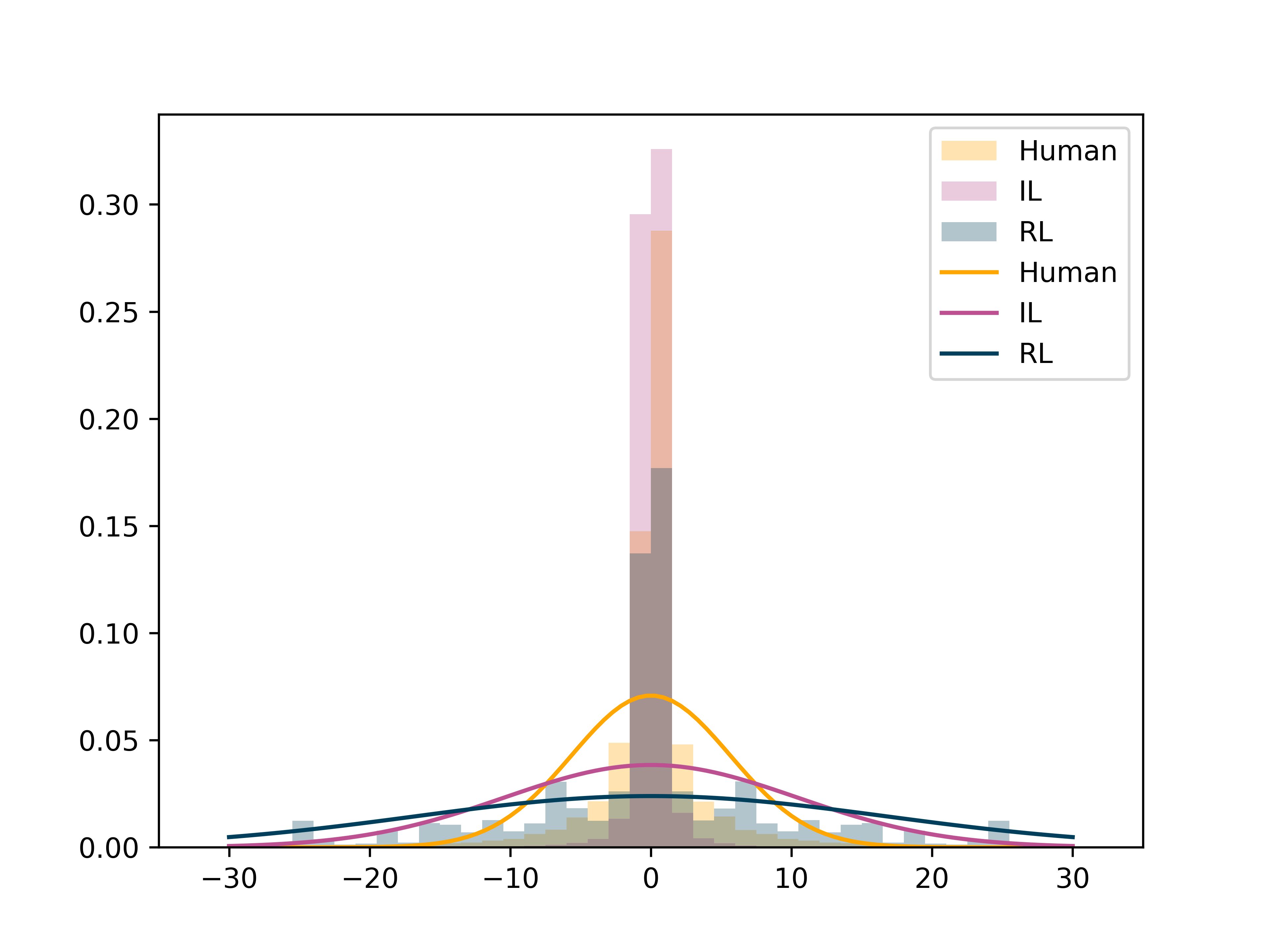}
    \caption{Histogram and Gaussian approximations comparing a) angular velocity b) angular acceleration, c) angular jerk, behaviours for camera movements} \label{fig:histograms}
\end{figure}

In this section we analyse the camera movement of the agents. The agent moves the camera to aim at enemies, navigate around the environment, or look around. In this analysis we look at three orders of movement. Order 1 corresponds to the change in camera motion per frame (time) analogous to angular velocity. Similarly order 2 captures angular acceleration and order 3 captures angular jerk. The camera movement behaviour is aggregated into a histogram to summarise the preferences. Figure \ref{fig:histograms} shows the histograms for comparing camera movements for Human, IL agent, and RL agent. Figure \ref{fig:histograms} also shows approximated Gaussian distributions modelled from behaviour data. Finally, these behaviours are quantitatively compared using a Wasserstein distance that compares the pairwise difference in distributions, in Table \ref{agent_perf}.

In Figure \ref{fig:histograms} it can be observed that Humans and IL agents both have a centre heavy distribution showing that they prefer no movement or slower movement if possible and execute extreme actions very infrequently, potentially only when required. RL agents however illustrate contrasting behaviour. RL agents have higher proportion of quick movements shown as a flattened Gaussian distribution with high variance and high frequency values at extreme angular velocities. Humans are expected to have smooth movements and less jerky motion as characterised in previous studies \cite{HNTT2022} often attributed to energy conservation behaviour \cite{energyBias2021}. Our empirical results show similar contrasting trends between Humans and RL agents.

Also, to note the RL agents follow binned discrete actions across a limited range for angular velocity and hence impose artificial constraints. A peak at the extreme of the distribution show that agents prefer fast motion. Hence, the agent might attempt to leverage even higher velocity magnitudes if the maximum velocity magnitude is changed. The peak around the extremes (-15 and 15) are hence artificial. The high values provide a competitive advantage by quick aiming and fast reactions.

Further, extending this analysis to higher order of camera movements (angular acceleration and angular jerk) we observe analogous dissimilarities between humans and RL agents. The human behaviour is heavily biased around zero corresponding to slower and smoother motion. RL agents have much higher values of angular acceleration and jerk. This variance is further exaggerated at higher order due to no direct control or constrained limit to maximum values. One critical difference from angular velocity behaviour is a major difference in behaviour between Human and IL agents. The difference between them is further highlighted in approximated Gaussians. IL agents are more varied and less centre heavy. Potential cause of this is that there is no direct signal while training for learning higher order motion.

 Wassertein distance is used to quantitatively compare agents and humans and the results are shown in Table \ref{tab:agent_perf}. The Wassertein distance employs a cumulative distribution function to quantify the differences in camera movement behaviour. This acts as a confirmation for the observations made from the histograms. For angular velocity there is a comparatively small difference between Human and IL agents. While both Human, RL and IL, RL have larger differences. This shows a successful replication of behaviour by IL agent in terms of camera movement. Further it provides insights into contrasting high performing behaviours. Both humans and RL agents achieve high number of kills but have different approaches as shown by different histogram distributions. For higher order comparisons the differences between IL agents and RL agents are not as large however, Human IL agents still maintain the smallest distances comparatively. This illustrates the current limitation with the two learning processes and motivates a need for further research to get closer to human-like behaviours.

\begin{table}[ht]
\begin{tabular}{lrrr}

\textbf{Wasserstein Distance}                         & \multicolumn{1}{l}{\textbf{Velocity}} & \multicolumn{1}{l}{\textbf{Acceleration}} & \multicolumn{1}{l}{\textbf{Jerk}} \\ \hline

\multicolumn{1}{l|}{Humans vs IL agent} & 0.0214 &0.0182 & 0.0261 \\
\multicolumn{1}{l|}{Humans vs RL agent} & 0.1407   &0.0243 &  0.0568 \\
\multicolumn{1}{l|}{IL agent vs RL agent} & 0.1359  &0.0405 &  0.0789 \\ \hline
\end{tabular}
\caption{Pairwise Wasserstein distances}
\label{agent_perf}
\end{table}

\section{Discussion and Conclusion}
\label{sec:Discussion}

This paper explored the use of pixel data and associated actions to train agents that learn via behavioural cloning, and reinforcement learning. The best-performing trained agent explored and navigated the map very well, with similar heat map patterns to the best player in our dataset. A large amount of the agent's time was spent quickly navigating to the centre and larger corridors where the majority of the action took place. 

The actual performance of the agent was roughly on par with the average human player, with the best agent achieving on average 11 kills and 1463 damage in three minutes. This puts the IL trained agent at a much higher level than the lowest performing human players in our data.

The individually trained agents perform slightly worse than their human training data, and slightly worse than the best agent, but have the added benefit of sharing inherent behavioural traits of their human data - such as navigation, and in-game reaction and mouse movement patterns.

The work in this paper outlines a method of training agents in games without the complex necessity for game engine data, by only capturing game frames and actions this type of data could be captured on any type of game and system. Although we also gathered depth data from the game, these can be generated using ML based depth estimators. Although not shown in this paper, we did find very similar results when using off-the-shelf the MiDaS depth estimator \cite{midas}.

On top of this, we show that multiple agent characteristics can be produced by only using simple pixel data, with the agent's navigation and in-game performance varying quite observably between trained agents.

We show an empirical comparison of the behaviours generated by Human, IL agent, and RL agents. We observe similarities in spatial preferences as observed in the heatmap between Humans, IL agents, and RL agents. While there are similarities in motion, the camera movements show the gap identifying a facet of human-like behaviour that the RL agents do not naturally acquire. It is observed that Humans tend to prefer slow, smooth, and energy conserving movements while controlling the camera and performing aiming movement. IL agents are trained to replicate this and hence behave similar to Humans. In contrast RL agents only optimise for objectives like kills, and hence exploit fast reflexes and jerky behaviour to achieve high performance. As a whole, this analysis highlights the differences in generated behaviours and underscores the need for IL learning metrics in RL agents to maintain human-like mechanisms of movements while optimising for performance objectives.

Long-sequence learning in ML remains a challenge, so we offer several improvements on the typical network and data-loading techniques. Exponential frame skipping offers a much larger view window for the network, without increasing the memory consumption through a linear, but larger window. Using a signed mask for the regressive loss of sign-sensitive predictions like that of the mouse offered a much more human-like movement from the agent, with smoother and less stepped control of the camera. Lastly, we show that using an average of several frames for the target label, over just using one frame for the target, offered much more stability in training through smoother labels of overlapping frames.

{\small
\bibliographystyle{ieee_fullname}
\bibliography{imitation}
}

\end{document}